\def\year{2019}\relax
\documentclass[letterpaper,twocolumn,10pt, conference]{ieeeconf}
\pdfoutput=1

\IEEEoverridecommandlockouts
\usepackage{helvet}  
\usepackage{courier}  
\usepackage{url}  
\usepackage{graphicx}  

\usepackage{amssymb,amsfonts,amsmath,mathrsfs,amsthm,bm}
\usepackage{mathtools}
\usepackage{latexsym}
\usepackage{algorithm}
\usepackage{algpseudocode}
\usepackage{epsfig,psfrag,color,graphics,epsf}
\usepackage{enumerate}
\usepackage{breqn}
\usepackage{float}

\DeclarePairedDelimiter\norm{\lVert}{\rVert}

\newtheorem{theorem}{\textbf{Theorem}}
\newtheorem{proposition}{\textbf{Proposition}}
\newtheorem{remark}{\textbf{Remark}}
\newtheorem{assumption}{\textbf{Assumption}}

\begin{document}

\title{Distributed SGD Generalizes Well Under Asynchrony}
\author{Jayanth Regatti \and Gaurav Tendolkar \and Yi Zhou \and Abhishek Gupta \and Yingbin Liang
\thanks{All authors are with ECE department at The Ohio State University, Columbus, OH. Email: {\tt \small regati.1@osu.edu, tendolkar.3@osu.edu, zhou.1172@osu.edu, gupta.706@osu.edu, liang.889@osu.edu}. J. Regatti and A. Gupta gratefully acknowledge ARPA-E NEXTCAR program for supporting this research. The work of Y. Liang was partially supported by the U.S. National Science Foundation under the grants CCF-1761506 and CCF-1900145. Results presented in this paper were obtained using Ohio Super Computer Center\cite{OhioSupercomputerCenter1987}, and the Chameleon testbed supported by the National Science Foundation.}
}

\maketitle

\begin{abstract}
The performance of fully synchronized distributed systems has faced a bottleneck due to the big data trend, under which asynchronous distributed systems are becoming a major popularity due to their powerful scalability. In this paper, we study the generalization performance of stochastic gradient descent (SGD) on a distributed asynchronous system. The system consists of multiple worker machines that compute stochastic gradients which are further sent to and aggregated on a common parameter server to update the variables, and the communication in the system suffers from possible delays. Under the algorithm stability framework, we prove that distributed asynchronous SGD generalizes well given enough data samples in the training optimization. In particular, our results suggest to reduce the learning rate as we allow more asynchrony in the distributed system. Such adaptive learning rate strategy improves the stability of the distributed algorithm and reduces the corresponding generalization error. Then, we confirm our theoretical findings via numerical experiments.

\end{abstract}

\section{Introduction}\label{sec:intro}
    Stochastic gradient descent (SGD) and its variants (e.g., Adagrad, Adam, etc) have been very effective in solving many challenging machine learning problems such as training deep neural networks. In practice, the solution found by SGD via solving an empirical risk minimization problem typically has good generalization performance on the test dataset. Recently, there has been a series of theoretical studies that establish generalization error bounds for SGD in nonconvex optimization \cite{hardt2015train,zhou2018generalization,Kuzborskij_2017}. They show that SGD can generalize well given enough training data samples, justifying in part its practical success.

   As the data volume in machine learning applications grows, traditional SGD cannot handle big data problems due to its sequential nature. Thus, various works have proposed distributed implementations of SGD, where multiple workers collaborate together to speed up the convergence time of SGD while maintaining its desirable convergence rate \cite{agarwal2011distributed}, \cite{zinkevich2010parallelized}. However, the overall performance of the distributed system is bottlenecked by full synchronization overhead in practical scenarios.  In specific, straggler workers and possible communication delays among the workers can significantly slowdown the convergence under full synchronization protocol. To deal with these issues, asynchronous protocol has been introduced to reduce the synchronization overhead of distributed systems. Such an asynchronous protocol has led to various kinds of distributed asynchronous SGD \cite{recht2011hogwild}, \cite{dean2012large}, \cite{liu2015asynchronous}, and the corresponding convergence rate is comparable to that of its full-synchronization counterpart.

While the effect of asynchrony on the convergence of distributed SGD has been extensively studied \cite{recht2011hogwild}, \cite{zhang2015staleness}, \cite{dutta2018slow}, whether the converged solution generalizes well on the testing data (i.e., unseen data samples) under asynchrony has not been explored. The aim here is to use the stability based framework to analyze the generalization performance of the distributed asynchronous SGD. In contrast to the existing such studies of SGD
\cite{hardt2015train,zhou2018generalization,Kuzborskij_2017}, the major challenge to analyze distributed SGD is due to the asynchrony so that
the gradients sent by the workers are stale (computed using the variables of a previous instance) and random in nature. Hence, the analysis of the stability bound requires new and sophisticated technical development of the iteration properties. To the best of the authors' knowledge, this is the first guaranteed analysis of the generalization error for the distributed asynchronous SGD. We summarize our contribution as follows.

\subsection{Our Contributions}
We study the generalization error of distributed asynchronous SGD, whose main update is performed on a parameter server that aggregates possibly delayed stochastic gradients computed by multiple worker machines, where the maximum delay is bounded by $\bar{\tau}$. To our best knowledge, this is the first study of the generalization error under the asynchronous SGD protocol \textcolor{black}{for nonconvex functions}.

To be specific, under the algorithm stability framework \cite{bousquet2002stability}, we establish a bound for the expected generalization error of distributed asynchronous SGD in nonconvex optimization. Our generalization bound shows that there is a degradation of generalization error in the presence of large delays however the error can be controlled provided large enough training data samples and a carefully chosen learning rate. 
Such theoretical result is further confirmed in our numerical experiments. Thus, under such choice of learning rate, the convergence of distributed asynchronous SGD is guaranteed (e.g., \cite{zhang2015staleness}), while the system benefits from speed up induced by parallelization and low generalization error due to high algorithm stability.

\subsection{Related Work}\label{sec.relatedwork}
\textbf{Distributed asynchronous SGD:}
The study of asynchronous algorithms dates back to the works \cite{tsitsiklis1986distributed,bertsekas1989parallel,Baudet78}. Such type of algorithms has attracted further attention in many recent works \cite{delay_pga_linear,Tseng_linear_PAA}.

Distributed asynchronous gradient-based algorithms have been studied in \cite{bertsekas1989parallel,Zhou2016,zhou2018distributed} under model parallelism and in \cite{agarwal2011distributed,ssp_parameter,parameter_server} under data parallelism, respectively.
These works show convergence and robustness of GD and SGD in the presence of stale gradient updates computed across several machines in a master-worker setting. \cite{zinkevich2010parallelized} present a parallelized stochastic gradient descent (synchronous) and provide a detailed analysis and experimental evidence. They use contractive mappings to quantify the speed of convergence of parameter distributions to their asymptotic limits.
\cite{graphlab}, \cite{agarwal2011distributed} study the convergence of gradient descent algorithms that use delayed gradients. They base their analysis on specific architectures (cyclic delayed architecture, locally averaged delayed architecture) and showed that optimization error of $n-$node architectures scales asymptotically as $\mathcal{O}(1/\sqrt{nT})$ after $T$ iterations. While the above two works are based on a synchronous setting, \cite{recht2011hogwild} present a novel update scheme (which they called HOGWILD!) that doesn't require memory locks and works in an asynchronous setting. They bounded the delay (of gradient update) variable by $\bar{\tau}$ and their convergence rate mimics that of serial SGD when $\bar{\tau} =0$.

\textbf{Generalization error of SGD:}
    The study of relationship between algorithms' stability and their corresponding generalization error is conducted by \cite{bousquet2002stability}, where they defined a notion of uniform stability that upper bounds the generalization error of symmetric and deterministic learning algorithms. This work is further extended to study the stability and generalization error of randomized learning algorithms in \cite{elisseeff2005stability}. The gist of their work is to show that a stable algorithm can generalize better. \cite{shalev2010learnability} developed various properties of stability on learning problems. In \cite{hardt2015train}, the authors first applied the stability framework to study the expected generalization error for SGD, and \cite{Kuzborskij_2017} further provided a data dependent generalization error bound. In \cite{Mou_2017}, the authors studied the generalization error of SGD with additive Gaussian noise. In \cite{Charles_2017}, the authors studied the generalization error of several first-order algorithms for loss functions satisfying the gradient dominance and the quadratic growth conditions. \cite{Poggio_2011} studied the stability of online learning algorithms. More recently, \cite{zhou2018generalization} establishes a variance-dependent generalization bound for SGD with probabilistic guarantee. While the above studies on generalization of SGD considered a serial case, \cite{yin2018gradient} studies the role that gradient diversity and \textcolor{black}{mini-batch size plays in characterizing the expected generalization error of mini-batch distributed SGD, which can analogously be looked at as synchronized distributed SGD. However, our setup is different than theirs with respect to the random sampling of the data used at each update. Moreover, their analysis does not hold for non-convex functions.}

\section{Problem Formulation and Preliminaries}\label{sec.problemformulation}

    \textcolor{black}{In this section, we describe the setup we consider for the analysis of distributed asynchronous SGD.}  We consider solving the following finite-sum optimization problem via distributed stochastic gradient decent (described in the next subsection):
    \begin{align}
        \min_{\mathbf{w} \in \mathbb{R}^d} F_n(\mathbf{w}):= \frac{1}{n} \sum_{i=1}^n f(\mathbf{w}; \mathbf{z}_i), \tag{P}
    \end{align}
    where function $f$ corresponds to a smooth and possibly non-convex loss and $\mathcal{S}:= \{\mathbf{z}_1, \ldots, \mathbf{z}_n \}$ denote $n$ training data samples that are drawn i.i.d from an underlying distribution $\mathcal{D}$.

\subsection{Distributed Asynchronous SGD}

    Consider a distributed system with $p$ workers, which are connected to a common parameter server. Divide the whole training dataset $\mathcal{S}$ into $p$ disjoint subsets $\{\mathcal{S}_j\}_{j=1}^p$ with equal cardinality, and distribute the $p$ subsets of data respectively to the $p$ workers. Each worker $j$ samples a data point from $\mathcal{S}_j$ uniformly at random. Then, the worker computes the gradient of the loss over the sampled data and sends it to the parameter server, where the stochastic gradients computed by the workers are aggregated and further applied to update the variables. After that, the parameter server sends the updated variables to the workers for computing gradients that are used in the next update. Assume that we initialize the variables across all workers as $\mathbf{w}_{0} \in \mathbb{R}^d$. Then, \textcolor{black}{in the perfectly synchronized case, }the update rule of distributed asynchronous SGD on the parameter server can be written as, for $t=0,...,T-1$,
    \begin{align*}
        \mathbf{w}_{t+1} = \mathbf{w}_{t} - \frac{\gamma_{t}}{p} \sum_{j=1}^{p} \nabla f(\mathbf{w}_t; \mathbf{z}_{\xi(j,t)}),
    \end{align*}

    where $\xi(j,t)$ is the index of the datapoint sampled by worker $j$ at time $t$, $\{\mathbf{w}_{t}\}_t$ denotes the variable sequence, $\gamma_t$ is the learning rate applied at the $t$-th update. It is easy to observe that the same update equation applies to the mini-batch setting of \cite{yin2018gradient}. The difference lies in the way the data points are sampled, in our case each data point is uniformly drawn from the dataset assigned to each worker, whereas in their case, each datapoint of the minibatch is uniformly drawn with replacement.

\begin{figure}
\centering
\includegraphics[width=3in]{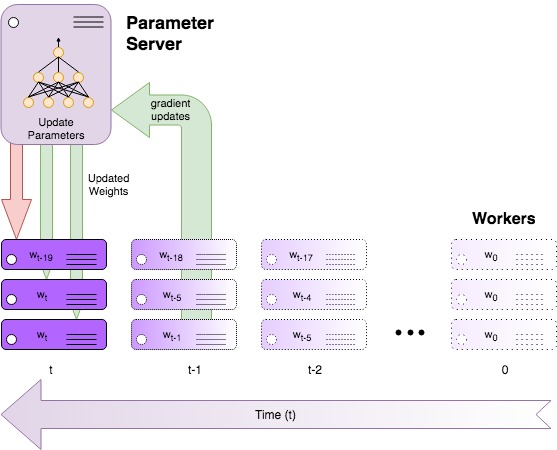}
\caption{Each worker machine has a copy of the model. At every time step, each worker computes a stochastic gradient using the local model. The parameter server aggregates the gradients from workers and updates the model via SGD. The updated model is then broadcast to all workers. The workers do not wait for the server to send fresh model, and compute gradients using the out-dated local model. The stochastic gradients received by the parameter server are also stale-synchronized.}
\end{figure}

In practical scenarios, achieving synchronization is difficult due to issues like  limited communication bandwidth between the workers and the parameter server, and straggling workers due to limited computation power or accidental shut downs. In such restrictive and undesirable scenarios, the whole system cannot be fully synchronized. As a consequence, the stochastic gradients sent from, say, worker $j$ to the parameter server at the $t$-th update can be a ``delayed" gradient that is computed based on an out-dated variable  $\mathbf{w}_{t-\tau(j,t)}$, where $\tau(j,t)$ denotes the corresponding delay. The update equation in this case is given by
    \begin{align}\label{eqn.delayedupdate}
        \mathbf{w}_{t+1} = \mathbf{w}_{t} - \frac{\gamma_{t}}{p} \sum_{j=1}^{p} \nabla f(\mathbf{w}_{t-\tau(j,t)}; \mathbf{z}_{\xi(j,t)})
    \end{align}

Note that we assume that the parameter server has access to the (delayed) stochastic gradients generated by all the workers at each update. \textcolor{black}{A practical case for this assumption is when computation is cheap and communication bandwidth is different for different workers, similar to the delayed architectures considered in \cite{agarwal2011distributed}. Throughout, we assume a fixed arbitrary delay sequence $\bm{\tau}:=\{\tau(1,t),...,\tau(p,t)\}_{t=1:T-1}$ where
$\tau(j,t)$ satisfies the following conditions, $0 \leq \tau(j,t) \leq \tau(j,t-1)+1 \leq \bar{\tau}$, where $\bar{\tau}$ is the maximum delay of the system. These conditions ensure that the staleness of the gradients can grow by a maximum of one at each time step. In the worst cases that a worker fails and sends the same gradient at each time step, the delay grows only by 1 at each update. Another simple example is when the delay is assumed constant. The output of the distributed asynchronous SGD is denoted by $\mathbf{w}_{T,\mathcal{S}}$, which depends on the data set $\mathcal{S}$, the sample path $\bm{\xi}:=\{\xi(1,t),...,\xi(p,t)\}_{t=1:T-1}$, and the delay path $\bm{\tau}$. In our analysis we consider the worst case delays, and hence our results hold for any delay path $\bm{\tau}$}.

\subsection{Stability and Generalization Error}\label{subsec.stab_gen}
In this subsection, we study the relationship between algorithmic stability of distributed asynchronous SGD and its generalization error.

Typically, the finite-sum optimization problem (P) is viewed as a sample-mean approximation of the following population risk minimization problem
\begin{align}
\min_{\mathbf{w} \in \mathbb{R}^d} F(\mathbf{w}):= \mathbb{E}_{\mathbf{z} \in \mathcal{D}}~ f(\mathbf{w}; \mathbf{z}), \tag{Q}
\end{align}
where we assume that the data is drawn from an underlying  distribution $\mathcal{D}$. The gap between the objective value of problem (P) and (Q) is defined as the generalization error, and we are interested in the expected generalization error, i.e.,
\begin{align*}
\text{(Generalization error):} \quad \Big|\mathbb{E}_{\mathcal{S},R} [F_n(\mathbf{w}_{\mathcal{S}}) - F(\mathbf{w}_{\mathcal{S}})]\Big|,
\end{align*}
where $\mathbf{w}_{\mathcal{S}}$ is output of the algorithm (dependent on data) and the expectation is taken over the draw of the dataset $\mathcal{S}$ and the \textcolor{black}{internal randomness $R$ of the algorithm (due to distributing $\mathcal{S}$ to the workers and $\bm{\xi}$)}. In particular, we are interested in the generalization error evaluated at the output of the distributed asynchronous SGD via solving the problem (P).

    The generalization error of the output of randomized learning algorithms has been studied under various theoretical frameworks, and algorithm stability is a popular one that has been recently applied to study the generalization error of SGD \cite{hardt2015train} and distributed synchronous SGD \cite{yin2018gradient}. We adopt the same notion of stability as in \cite{hardt2015train} as follows. Suppose $\mathcal{S}, \mathcal{S}'$ are two datasets that differ at a single data point, then $\epsilon_{\text{stab}}$ is
    \begin{align*}
        \sup_{z \sim \mathcal{D}}\mathbb{E}_{R}[ f(\mathbf{w}_{T,\mathcal{S}};\mathbf{z}) - f(\mathbf{w}_{T,\mathcal{S}'};\mathbf{z})]\leq \epsilon_{\text{stab}}
    \end{align*}
    where $\mathbf{w}_{T,\mathcal{S}}$ is the output of the algorithm trained on $\mathcal{S}$ for $T$ updates.
    Intuitively, the stability measures the function value gap evaluated at the outputs of the algorithm that are obtained by solving the problem (P) with two datasets that are different at one data sample. It has been shown in \cite{hardt2015train} that algorithm stability bounds the corresponding generalization error of the output of the algorithm, i.e.,

    \begin{align*}
        \Big|\mathbb{E}_{R,S} [F_n(\mathbf{w}_{T,S}) - F(\mathbf{w}_{T,S})] \Big|\leq \epsilon_{\textrm{stab}}
    \end{align*}

Thus, our goal is to study the stability of distributed asynchronous SGD.

\section{Stability of Distributed Asynchronous SGD in Nonconvex Optimization}
\label{sec.generalization}
To study stability of distributed asynchronous SGD,
we make the following standard assumptions regarding the loss function in problem (P).

\begin{assumption}\label{assum: f}
    For all $\mathbf{z} \sim \mathcal{D}$, the loss function $f$ satisfies:
    \begin{enumerate}
        \item Function $f(\cdot ; \mathbf{z})$ is continuously differentiable;
        \item Function $f(\cdot ; \mathbf{z})$ is non-negative and $L$-Lipschitz continuous; The norm of $\nabla f$ is uniformly bounded by $L$, i.e. $\sup_{\mathbf{z}}\norm{\nabla f(\cdot;\mathbf{z})} \leq L$;
        \item Function $f(\cdot ; \mathbf{z})$ is $\beta$-smooth,  i.e., $\nabla f(\cdot;\mathbf{z})$ is $\beta$-Lipschitz continuous.
    \end{enumerate}
\end{assumption}

{
\begin{remark}
Although the norm of gradient can be unbounded in the entire space, it can be controlled via stability-inducing operations, e.g., regularization, projection and gradient clipping, etc. as discussed in \cite{hardt2015train}.
\end{remark}
}

In our setting, we consider two datasets $\mathcal{S}=\{\mathbf{z}_{1}, \mathbf{z}_{2},...,\mathbf{z}_{n}\}$, and $\bar{\mathcal{S}}=\{\mathbf{z}'_{1}, \mathbf{z}'_{2},...,\mathbf{z}'_{n} \}$ that are independent random samples from $\mathcal{D}$. Let $\mathcal{S}^{'} = \{\mathbf{z}_{1},..., \mathbf{z}'_{i},...,\mathbf{z}_{n}\}$ be the sample that is identical to $\mathcal{S}$ except for the $i$-th data sample. Let $I$ be the event that the two datasets $\mathcal{S}$ and $\mathcal{S}^'$ differ at the $i^{*}$-th data point. Let $J$ be the event that the data is uniformly divided into $p$ chunks and that the $i^{*}$-th data sample belongs to the $j^{*}$-th worker. Lastly, denote $\mathbf{w}_{T,\mathcal{S}},\mathbf{w}_{T,\mathcal{S}^'}$ as the $T$-th outputs of the algorithm obtained by solving problem (P) with training datasets $\mathcal{S}$ and $\mathcal{S}^'$ respectively. The randomness $R$ of the algorithm is due to the randomness of $I, J, \bm{\xi}$. \textcolor{black}{By considering the sampling of data in this way (as compared to with replacement sampling of \cite{yin2018gradient}, we avoid the case where we encounter the different sample multiple times at the same update, thus simplifying the stability analysis. Note that we still need to deal with the staleness of the gradients.}

From the Lipschitz property of $f$, we can write
\begin{align}
       \mathbb{E}_{R}[ f(\mathbf{w}_{T,\mathcal{S}};\mathbf{z}) - f(\mathbf{w}_{T,\mathcal{S}'};\mathbf{z})] \leq L \mathbb{E}_{R}\norm{\mathbf{w}_{T,S} - \mathbf{w}_{T,S^{'}}}. \label{eq: 11}
   \end{align}

We define $\delta_{t}$ as the norm of the difference between the parameter vectors $\mathbf{w}_{t,\mathcal{S}}$ and $\mathbf{w}_{t,\mathcal{S}^{'}}$ of the algorithm run on data sets $\mathcal{S}$ and $\mathcal{S}^{'}$ respectively. For notational simplicity, we denote $\mathbf{w}_{t,\mathcal{S}}$ by $\mathbf{w}_{t}$ and $\mathbf{w}_{t,\mathcal{S}^{'}}$ by $\mathbf{\bar{w}}_{t}$ for any $0<t\leq T$. Therefore, we can write $\delta_{t} = \big|\big|\mathbf{w}_{t}-\mathbf{\bar{w}}_{t}\big|\big|$. Our first result is a bound on the divergence of $\delta_{t}$ as the algorithm iterates from $t=1,...,T$, presented as the following proposition.
\begin{proposition}\label{lem.2}
    Let Assumption \ref{assum: f} hold and run the distributed asynchronous SGD on two datasets $\mathcal{S}$, and $\mathcal{S}^{'}$ that differ at a single data sample. Denote the generated sequences of variables as $\{\mathbf{w}_{t}\}_t, \{\Bar{\mathbf{w}}_{t}\}_t$ respectively, denote $\delta_{t} = \norm{\mathbf{w}_{t}-\Bar{\mathbf{w}}_{t}}$, and $\delta_{t}=0$. Then, $\mathbb{E}_{R}[\delta_{t}]$ satisfies the following recursion
    \begin{equation}
        \mathbb{E}_{R}[\delta_{t+1}] \leq   \mathbb{E}_{R}[\delta_{t}] +
        \underbrace{\beta \gamma_{t}(\bar{\tau}+1) \max_{t-\bar{\tau} \leq k \leq t}\mathbb{E}_{R}[\delta_{k}]}_{A} +  \underbrace{\frac{2L\gamma_{t}}{n}}_{B}. \nonumber
    \end{equation}
\end{proposition}

Proposition \ref{lem.2} establishes a recursive property of $\mathbb{E}_{R} [\delta_t]$, which measures the expected stability of the iterate variables generated by distributed asynchronous SGD. The recursion characterizes the effect of maximum delay $\bar{\tau}$ on the stability. \textcolor{black}{The term $A$ follows from the presence of delay and the difference between the parameters $\delta_k$ at previous time steps, and $B$ follows from the different sample. }

It can be seen that a larger $\bar{\tau}$ blows up the stability bound, making the distributed algorithm more {\em unstable}, which is consistent with one's intuition, because delayed information introduces more turbulence into the system. On the other hand, one natural way to mitigate the negative effect of delay is to reduce the learning rate $\gamma_t$ in the recursion. In fact, reducing the learning rate \textcolor{black}{(inversely proportional to delays)} has been proven to guarantee the convergence of asynchronous SGD \cite{recht2011hogwild}. Our next result establishes that it can also help to stabilize the algorithm and improve its generalization performance.

We further elaborate the choice of learning rate in the main theorem later. Next, we establish the following useful proposition in order to telescope the recursion in Proposition \ref{lem.2} for deriving our main result. The proof is presented in the appendix.

\begin{proposition}\label{lem.induc}
    Let $V(t)$ be a sequence of real numbers satisfying, for $\ t=0,1,2,...$
    \begin{equation}\label{eqn.lemrec}
        V(t+1) \leq V(t) + q_t \max_{t-\tau(t) \leq s \leq t} V(s) + r_{t},
    \end{equation}
    for some non-negative numbers $q_{t}$ and $r_{t}$. If $V(0)=0$, $ 0 \leq \tau(t) \leq \bar{\tau},$
    then
    \begin{equation}\label{eqn.induction}
        V(T+1) \leq \sum_{t=0}^{T} \bigg( \prod_{k=t}^{T} (1 + q_{k}) \bigg) r_{t}
    \end{equation}
\end{proposition}

Note that the iteration-dependent terms $q_t$ and $r_t$ captures a key difference of Proposition~\ref{lem.induc} from Lemma 3 in \cite{feyzmahdavian2014delayed}, in which $q_t$ and $r_t$ become universal constants that are independent of $t$. Moreover, our proof of Proposition~\ref{lem.induc} requires the construction of the structure in \eqref{eqn.induction} for induction. Such a structure and the resulting inductive argument are very different and more challenging than those in \cite[Lemma 3]{feyzmahdavian2014delayed}.



Next, we apply Proposition \ref{lem.induc} to the recursion in Proposition \ref{lem.2} and obtain our main result on stability (or equivalently, generalization error) of distributed asynchronous SGD.
\begin{theorem}[Stability bound]\label{thm.1}
Let Assumption \ref{assum: f} hold and assume that the maximum delay of the distributed system is bounded by $\bar{\tau} \in \mathbb{N}$. Apply distributed asynchronous SGD for $T$ updates to solve problem (P) and choose learning rate
\textcolor{black}{$\gamma_{t} \leq \frac{c}{(t+3)}$, where $c>0$ is an arbitrary constant. Then, the stability of the algorithm is bounded by
$$
\epsilon_{\textrm{stab}} \leq \frac{2L^2(T+3)^{\beta c(\bar{\tau}+1)}}{n\beta (\bar{\tau}+1)}.
$$
}
\end{theorem}
The proof of Theorem \ref{thm.1} is presented in the appendix. In particular, our proof needs to handle the effects caused by delayed stochastic gradients and variables in the distributed system, as opposed to \textcolor{black}{\cite{hardt2015train} (serial), and \cite{yin2018gradient} (mini-batch/fully synchronized).}

\textcolor{black}{Theorem \ref{thm.1} establishes the stability bound for distributed asynchronous SGD after $T$ updates. It can be seen that the algorithm stability vanishes sublinearly as the total number of training samples $n$ goes to infinity, meeting the dependence on $n$ in existing stability bounds for nonconvex SGD \cite{hardt2015train,Kuzborskij_2017}. Thus, distributed asynchronous SGD can generalize well given enough training data samples and a proper choice of the stepsize. In the special case of full synchronization, i.e., $\tau=0$, the bound in Theorem~\ref{thm.1} is similar to that in \cite{hardt2015train}(which was developed for the serial case)}.

\textcolor{black}{Theorem \ref{thm.1} shows that  for large delays initially (when $T$ is small), the denominator dominates and the generalization error is small, but as $T$ grows, the polynomial term dominates and generalization error grows rapidly. However the tuneable parameter $c$ in the learning rate can be adjusted so as to compensate the inconsistency caused by a large delay (and hence the polynomial term). This also agrees the with the result of \cite{hardt2015train}, to train faster inorder to generalize better. An inverse dependence of the learning rate on the delay has also been adopted in the study of convergence guarantee of asynchronous SGD \cite{recht2011hogwild,zhang2015staleness}.
}

\textcolor{black}{Note that, as discussed in \cite{hardt2015train}, by randomly sampling the data points, the number of parameter updates before the algorithm encounters different data samples in $\mathcal{S}$ and $\mathcal{S}^'$ is quite large and until then, $\delta_t = 0$. Following a similar approach as Lemma 3.11 and Theorem 3.12 of \cite{hardt2015train}, we get the following result
}

\begin{theorem}\label{thm: 2}
\textcolor{black}{
Let $f(\cdot, z) \in [0,1]$ and assume all the conditions in Theorem \ref{thm.1} hold, we can improve the stability bound as
\begin{align*}
    \epsilon_{\text{stab}} &\leq \frac{p+\frac{p^{1/(k+1)}}{k}}{n}\bigg(2L^2c\bigg)^{\frac{1}{k+1}}\bigg(T+3\bigg)^{\frac{k}{k+1}}
\end{align*}
where $k=\beta c (\bar{\tau} + 1)$
}
\end{theorem}

\section{Experimental Evaluation}

In this section, we present experimental results to support our theoretical findings. Specifically, we show how maximum allowed delay affects the generalization error. We also observe that the impact of the delay on generalization error can be negated by carefully selecting the learning rate. Our experimental set up is as follows.

\begin{figure*}[!t]
    \centering
    \includegraphics[width=0.49\linewidth]{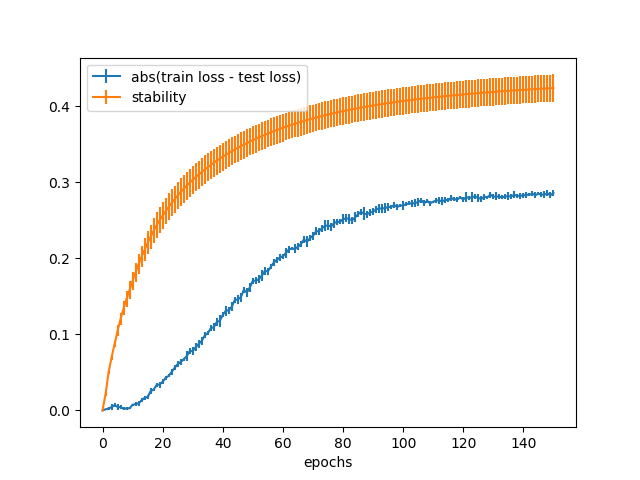}
    \includegraphics[width=0.49\linewidth]{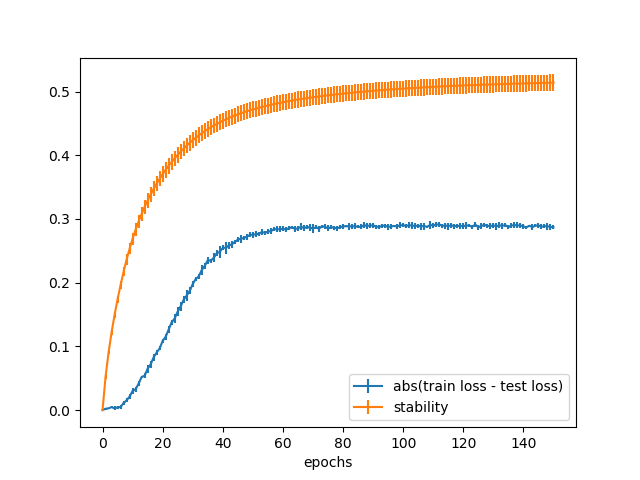}
    \caption{\label{fig: stab} Normalized Euclidean Distance between the parameters vs Generalization error measured for $\bar{\tau}=16$ (a) lr=$\frac{0.2}{(1 + 0.05 * t)}$, (b) lr=$\frac{0.4}{(1 + 0.05 * t)}$}
\end{figure*}

\begin{figure*}[!t]
    \centering
    \includegraphics[width=0.49\linewidth]{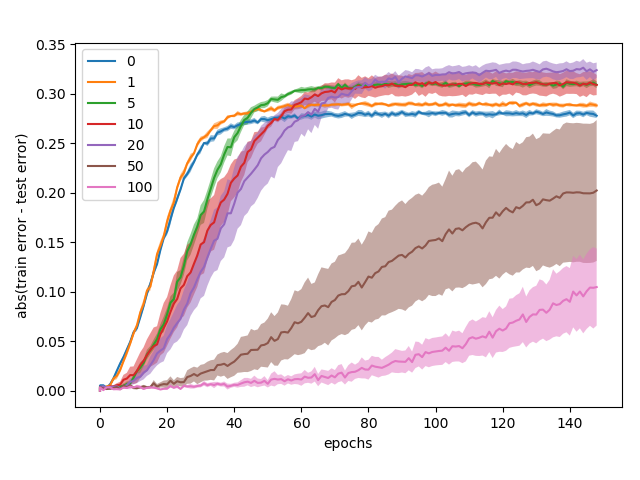}
    \includegraphics[width=0.49\linewidth]{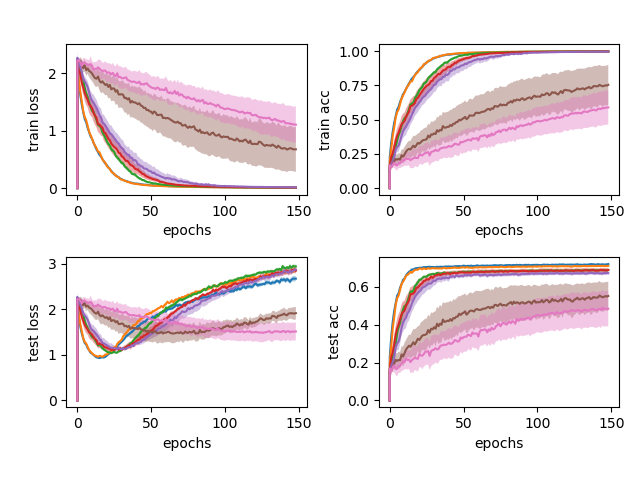}
    \caption{\label{fig: gen delay}Generalization error, training and testing plots for $\gamma_t=\frac{0.5}{(1+0.05*t)}$, and various values of $\bar{\tau}$. For large delays, training is slower, and overfitting starts after a long time (with a worse loss). After sometime however, generalization error starts to grow rapidly.}
\end{figure*}

\begin{figure*}
    \centering
    \includegraphics[width=0.49\linewidth]{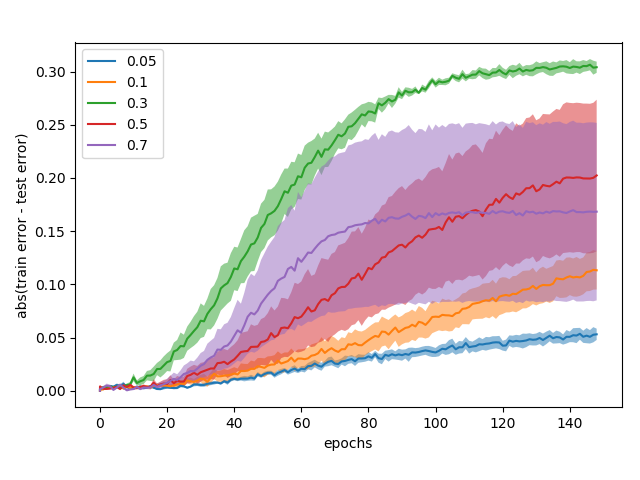}
    \includegraphics[width=0.49\linewidth]{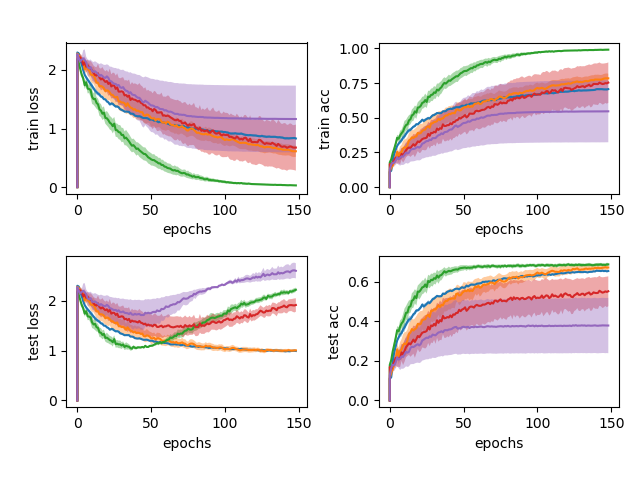}
    \caption{\label{fig: gen lr}Generalization error, training and testing plots for $\bar{\tau}=50$ and various values of learning rate coefficient $c$. In the presence of large delays, small learning rates lead to better generalization with comparable test accuracy, as compared to high learning rates. Note that the variance in the plots is due to choosing different delays (bounded by $\bar{\tau}$) in multiple trials.}
\end{figure*}

\textcolor{black}{
We train a VGGnet like model on CIFAR10 using distributed asynchronous SGD. In specific, the system consists of 8 workers that are connected to a common parameter server, and the dataset is distributed uniformly and equally to the workers. We use a batch size of 64 per worker and the gradient updates are performed using (\ref{eqn.delayedupdate}). We assign each worker with a fixed delay between $0$ and $\bar{\tau}$ (atleast one worker has $\bar{\tau}$ and one worker has no delay, this is done to ensure maximum possible staleness of gradients)}.

\textcolor{black}{We use a slightly different learning rate schedule than the one proposed in Theorem \ref{thm.1} as $\gamma_{t} = \frac{c}{1+0.05*t}$ (where $c$ is the learning rate coefficient), and the loss function is set to be cross entropy loss. We perform experiments with different values of maximum delay $\bar{\tau}$, learning rate coefficient $c$, number of workers $p$. We use the absolute value of the difference between the loss on the train and test data, and absolute value of the difference between the misclassification rate on the train and test data as proxies for measuring generalization error. We also plot the normalized Euclidean distance between the parameters using the formula $\sqrt{\norm{\mathbf{w}-\mathbf{w}^'}^2/(\norm{\mathbf{w}}^2 + \norm{\mathbf{w}^'}^2)}$ as a measure of stability of the algorithm. Note that since the goal here is to illustrate the effect of delays and learning rate, we did not perform hyperparameter tuning to choose the best learning rate or regularization to improve performance.}

\textcolor{black}{Figure \ref{fig: stab} shows that the normalized Euclidean distance between $\mathbf{w}, \mathbf{w}^'$ grows with the number of updates, and bounds the generalization error from above. From Figure \ref{fig: gen delay}, we can observe that if the delay is large, generalization error is lower in the beginning (delay in denominator dominates) but grows faster as $T$ increases (polynomial term dominates), and this transition takes longer when the learning rate used is larger. Note that the generalization error for large delays is lesser in the beginning due to large optimization error. We can observe that when the delay is small, the models start overfitting faster but end up with better models as compared to the case of large delays where the overfitting happens much later, but with poorer models. Figure \ref{fig: gen lr} shows that a smaller learning rate leads to a smaller generalization error holds true for small delays, but when the delays are large, even higher learning rate has a smaller generalization error (due to large training error) when $T$ is small, but grows fast as $T$ increases. In general, the presence of delays degrades the performance, however the choice of learning rate helps control the generalization error. Smaller learning rates leads to better performance when there are large delays in the system, and they achieve comparable model accuracy with negligible generalization error as compared to higher learning rates. The tradeoff, however, is the rate at which it converges.
}

\section{Conclusion}
In this paper, we presented the generalization error bound of distributed asynchronous SGD for a general non convex problem using algorithmic stability framework. \textcolor{black}{In the case that delay is zero, our bounds mimic that of the serial case}. Our bounds also explain the affect of maximum delay and learning rate coefficient on the generalization error, and corroborated our results with experimental data. Our future work includes developing high probability bounds for generalization in the asynchronous case. We believe that studying generalization of distributed asynchronous SGD is helpful in devising better algorithms that enjoy the benefits of parallel optimization while not compromising with generalization.

\appendices
\section{Proofs of Main Results}
We first prove two propositions and then proceed to prove Theorem \ref{thm.1}, which is our main result.
\subsection{Proof of Proposition \ref{lem.2}}
        Let $\mathcal{S}$ and $\mathcal{S^{'}}$ be two samples of size $n$ differing in only a single example, at index $i^{*} \in \{1,...,n\}$. Consider the case when the sample picked by processor $j^{*}$ at time $t$ is not $z_{i^{*}}$. This happens with probability $1-\frac{1}{n/p}$. Then in this case, we can find $\delta_{t+1}$ as
    \begin{align*}
    &\norm{\mathbf{w}_{t+1}-\Bar{\mathbf{w}}_{t+1}} \nonumber\\
    &= \Big\|\mathbf{w}_{t} - \frac{\gamma_{t}}{p} \sum_{j=1}^{p}\nabla f(\mathbf{w}_{t-\tau(j,t)}; \mathbf{z}_{\xi(j,t)})\\
    &\qquad - \Bar{\mathbf{w}_{t}} + \frac{\gamma_{t}}{p} \sum_{j=1}^{p}\nabla f(\Bar{\mathbf{w}}_{t-\tau(j,t)}; \mathbf{z}_{\xi(j,t)})\Big\|\\
    &\leq \norm{\mathbf{w}_{t}-\Bar{\mathbf{w}}_{t}} + \frac{\gamma_{t}}{p} \sum_{j=1}^{p}\norm{ \nabla f(\mathbf{w}_{t-\tau(j,t)}; \mathbf{z}_{\xi(j,t)}) \\
    &\qquad-\nabla f(\Bar{\mathbf{w}}_{t-\tau(j,t)}; \mathbf{z}_{\xi(j,t)})}\\
    &\leq \delta_{t} + \beta \frac{\gamma_{t}}{p} \sum_{j=1}^{p}\norm{\mathbf{w}_{t-\tau(j,t)} - \Bar{\mathbf{w}}_{t-\tau(j,t)}}
    \end{align*}
where the second inequality is due to the $\beta$-smoothness property of $f(\cdot;z)$.  This further yields
\begin{align*}
\delta_{t+1}&\leq \delta_{t} + \beta \frac{\gamma_{t}}{p} \sum_{j=1}^{p}\delta_{t-\tau(j,t)}.
\end{align*}
In the case that the processor $j^{*}$ picks sample $z_{i^{*}}$ (with probability $\frac{1}{n/p}$), then
    \begin{align}
        \begin{split}
                &\norm{\mathbf{w}_{t+1}-\Bar{\mathbf{w}}_{t+1}} \nonumber\\
                &=  \Big\|\mathbf{w}_{t} - \frac{\gamma_{t}}{p} \sum_{j=1}^{p}\nabla f(\mathbf{w}_{t-\tau(j,t)}; \mathbf{z}_{\xi(j,t)})\\
                &\qquad- \Bar{\mathbf{w}_{t}} + \frac{\gamma_{t}}{p} \sum_{j=1}^{p}\nabla f(\Bar{\mathbf{w}}_{t-\tau(j,t)}; \mathbf{z}_{\xi(j,t)})\Big\|\\
                &\leq \norm{\mathbf{w}_{t}-\Bar{\mathbf{w}}_{t}} + \frac{\gamma_{t}}{p} \sum_{\substack{j=1\\j\neq j^{*}}}^{p}\Big\|\nabla f(\mathbf{w}_{t-\tau(j,t)}; \mathbf{z}_{\xi(j,t)})\\
                &\quad-\nabla f(\Bar{\mathbf{w}}_{t-\tau(j,t)}; \mathbf{z}_{\xi(j,t)})\Big\|\\
                &\quad+ \frac{\gamma_{t}}{p}\Big\|\nabla f(\mathbf{w}_{t-\tau(j^{*},t)}; \mathbf{z}_{i^{*}}) -\nabla f(\Bar{\mathbf{w}}_{t-\tau(j^{*},t)}; \mathbf{z}_{i^{*}})\Big\|\\
                &\leq \delta_{t} + \beta \frac{\gamma_{t}}{p} \sum_{\substack{j=1\\j\neq j^{*}}}^{p}\norm{\mathbf{w}_{t-\tau(j,t)} - \Bar{\mathbf{w}}_{t-\tau(j,t)}} + \frac{2L\gamma_{t}}{p} \\
                &\leq \delta_{t} + \beta \frac{\gamma_{t}}{p} \sum_{j=1}^{p}\delta_{t-\tau(j,t)}+ \frac{2L\gamma_{t}}{p}.
        \end{split}
    \end{align}
    where the last but one inequality follows from the $\beta$-smooth and $L$-Lipschitz property of $f(\cdot;z)$. Note that $\delta_{t}\geq 0$ for all $t$.

    Now, taking expectation of $\delta_{t+1}$ with respect to the randomness of the algorithm, we get
    \begin{align}
            \mathbb{E}_{R}[\delta_{t+1}] \leq&\ \bigg(1-\frac{1}{n/p}\bigg) \mathbb{E}_{R}\bigg[\delta_{t} + \beta \frac{\gamma_{t}}{p} \sum_{j=1}^{p}\delta_{t-\tau(j,t)}\bigg] \nonumber\\
            &+ \frac{1}{n/p} \mathbb{E}_{R}\bigg[\delta_{t} + \beta \frac{\gamma_{t}}{p} \sum_{j=1}^{p}\delta_{t-\tau(j,t)}+ \frac{2L\gamma_{t}}{p} \bigg] \nonumber\\
            =&\ \mathbb{E}_{R}[\delta_{t}] + \beta \frac{\gamma_{t}}{p} \sum_{j=1}^{p}\mathbb{E}_{R}[\delta_{t-\tau(j,t)}] +  \frac{2L\gamma_{t}}{n}\nonumber\\
            \leq&\ \mathbb{E}_{R}[\delta_{t}] + \beta \frac{\gamma_{t}}{p} \sum_{j=1}^{p} \sum_{k=t-\bar{\tau}}^{t}\mathbb{E}_{R}[\delta_{k}] +  \frac{2L\gamma_{t}}{n}\nonumber\\
            =&\ \mathbb{E}_{R}[\delta_{t}] + \beta \gamma_{t} \sum_{k=t-\bar{\tau}}^{t}\mathbb{E}_{R}[\delta_{k}] +  \frac{2L\gamma_{t}}{n}\nonumber\\
            \leq&\ \mathbb{E}_{R}[\delta_{t}] + \beta \gamma_{t}(\bar{\tau}+1) \max_{t-\bar{\tau} \leq k \leq t}\mathbb{E}_{R}[\delta_{k}] +  \frac{2L\gamma_{t}}{n}. \label{eqn.reslem2}
    \end{align}
    where the second inequality is due to the fact that $\tau(j,t)\leq \bar{\tau}$, and the next equality follows because the inner summation doesn't depend on $j$ any longer.

\subsection{Proof of Proposition \ref{lem.induc}}
It is straightforward to verify that (\ref{eqn.induction}) holds true for $t=0,1$. Assume that it holds true for all   $\Bar{t} < T$, then
\begin{align}\label{eqn.initind}
    \begin{split}
        V(\Bar{t}+1) &\leq \sum_{t=0}^{\Bar{t}} \bigg( \prod_{k=t}^{\Bar{t}} (1+q_{k}) \bigg) r_{t},\ \ \ \Bar{t} < T
    \end{split}
\end{align}

Note that $\sum_{t=0}^{\Bar{t}} \bigg( \prod_{k=t}^{\Bar{t}} (1+q_{k}) \bigg) r_{t}$ is an increasing function in $\Bar{t}$. From (\ref{eqn.lemrec}) and (\ref{eqn.initind}), we have
\begin{align}
    \begin{split}
        V(T+1) \leq &\ V(T) +  q_{T} \max_{T-\tau(\cdot;T) \leq s \leq T} V(s) + r_{T}, \\
        \leq &\ \sum_{t=0}^{T-1} \bigg( \prod_{k=t}^{T-1} (1+q_{k}) \bigg) r_{t}\\ &+ q_{T} \sum_{t=0}^{T-1} \bigg( \prod_{k=t}^{T-1} (1+q_{k}) \bigg) r_{t} + r_{T} \\
        = &\ (1+q_{T}) \sum_{t=0}^{T-1} \bigg( \prod_{k=t}^{T-1} (1+q_{k}) \bigg) r_{t} + r_{T}\\
        = &\ \sum_{t=0}^{T-1} \bigg( \prod_{k=t}^{T} (1+q_{k}) \bigg) r_{t} +r_{T}\\
        = &\ \sum_{t=0}^{T} \bigg( \prod_{k=t}^{T} (1+q_{k}) \bigg) r_{t} + r_{T} - (1+q_{T})r_{T}\\
        \leq &\ \sum_{t=0}^{T} \bigg( \prod_{k=t}^{T} (1+q_{k}) \bigg) r_{t}
    \end{split}
\end{align}
where the final inequality follows from the fact that $1+q_{k} \geq 1$

\textcolor{black}{
Note that the result of Proposition\ \ref{lem.induc} can be modified for the case where $V(k)=0 $ for $0\leq k \leq t_0$ as follows
\begin{align}\label{eqn.extend prop}
    V(T+1) \leq \sum_{t=t_0}^{T}\bigg(\prod_{k=t}^{T}(1+q_k) \bigg)r_t
\end{align}
}

\subsection{Proof of Theorem \ref{thm.1}}

Define $V(t) = \mathbb{E}[\delta_{t}]$, $q_{t}= \beta \gamma_{t} (\bar{\tau} +1)$, and $r_{t}=\frac{2L\gamma_{t}}{n}$. Clearly, (\ref{eqn.reslem2}) satisfies the conditions for Proposition \ref{lem.induc}. Next, invoke Proposition~\ref{lem.induc} to obtain,
\begin{align*}
    \mathbb{E}[\delta_{T+1}] \leq &\ \sum_{t=0}^{T} \bigg( \prod_{k=t}^{T} (1+\beta \gamma_{k}(\bar{\tau}+1)) \bigg) \frac{2L\gamma_{t}}{n}\\
    \leq&\ \sum_{t=0}^{T} \bigg( \prod_{k=t}^{T} e^{\beta \gamma_{k}(\bar{\tau}+1)} \bigg) \frac{2L\gamma_{t}}{n}\\
    =&\ \sum_{t=0}^{T} \bigg(  e^{\beta (\bar{\tau}+1) \sum_{k=t}^{T}\gamma_{k}} \bigg) \frac{2L\gamma_{t}}{n}\\
    \leq&\ \textcolor{black}{\sum_{t=0}^{T} \bigg(  e^{\beta (\bar{\tau} +1)c\sum_{k=t}^{T}\frac{1}{k+3}} \bigg) \frac{2Lc}{n(t+3)}}\\
    \leq&\ \textcolor{black}{\sum_{t=0}^{T} \bigg(  e^{\beta(\bar{\tau}+1)c\log{\frac{T+3}{t+2}}} \bigg) \frac{2Lc}{n(t+3)}}\\
    \leq&\ \textcolor{black}{\frac{2Lc}{n} (T+3)^{\beta(\bar{\tau}+1)c} \sum_{t=0}^{T} (t+2)^{-\beta(\bar{\tau}+1)c -1}}  \\
    \leq&\ \textcolor{black}{\frac{2L}{n\beta(\bar{\tau}+1)} (T+3)^{\beta(\bar{\tau}+1)c} (1-(T+2)^{-\beta(\bar{\tau}+1)c})}\\
    \le&\ \textcolor{black}{\frac{2L(T+3)^{\beta(\bar{\tau}+1)c}}{n\beta (\bar{\tau}+1)}}
\end{align*}
where the second inequality uses $(1+x)\leq e^{x}$, third inequality uses the definition of $\gamma_{k}$, fourth inequality uses the fact that \textcolor{black}{$\sum_{t=t_{0}}^{T} \frac{1}{t+3} \leq \int_{t_0+2}^{T+3}\frac{1}{t}dt = \log{\frac{T+3}{t_{0}+2}}$, and the last but one inequality follows from the fact that $\sum_{t=0}^{T}(t+2)^{-c-1} \leq \int_{1}^{T+2}t^{-c-1} dt $}.  The desired stability bound follows from \eqref{eq: 11}.

\subsection{\textcolor{black}{Proof of Theorem \ref{thm: 2}}}

Let $\delta_{t_0} = 0$, then using (\ref{eqn.extend prop}), we get
\begin{align*}
    \mathbb{E}[\delta_{T+1}|\delta_{t_0}=0] \leq &\ \sum_{t=t_0}^{T} \bigg( \prod_{k=t}^{T} (1+\beta \gamma_{k}(\bar{\tau}+1)) \bigg) \frac{2L\gamma_{t}}{n}\\
\end{align*}
which leads to the following
\begin{align*}
    \mathbb{E}[\delta_{T+1}|\delta_{t_0}=0] \le&\ \frac{2L}{n\beta (\bar{\tau}+1)}\bigg(\frac{T+3}{t_0 + 2}\bigg)^{\beta c (\bar{\tau}+1)}
\end{align*}

Using the result of Lemma 3.11 \cite{hardt2015train}, we have for $t_0 \in \{1,...,n/p\}$ (since each worker has $n/p$ data points and we are only interested in the number of updates before the different sample is encountered)
\begin{align*}
    \mathbb{E}|f(\mathbf{w}_T; z) - f(\mathbf{w}^{'}_T;z)| \leq \frac{t_0}{n/p} + L\mathbb{E}[\delta_T|\delta_{t_0}=0]
\end{align*}

We then have by plugging in the above result,
\begin{align*}
    &\mathbb{E}|f(\mathbf{w}_T; z) - f(\mathbf{w}^{'}_T;z)| \\ &\qquad \leq \frac{t_0}{n/p} + \frac{2L^2}{n\beta (\bar{\tau}+1)}\bigg(\frac{T+3}{t_0 + 2}\bigg)^{\beta c (\bar{\tau}+1)}
\end{align*}

By minimizing with respect to $t_0$ and setting $k=\beta c (\bar{\tau} + 1)$, we get
\begin{align*}
    &\mathbb{E}|f(\mathbf{w}_T; z) - f(\mathbf{w}^{'}_T;z)| \leq \\ &\qquad\ \frac{p+\frac{p^{1/(k+1)}}{k}}{n}\bigg(2L^2c\bigg)^{\frac{1}{k+1}}\bigg(T+3\bigg)^{\frac{k}{k+1}}
\end{align*}
and since this holds for all $\mathcal{S}, \mathcal{S'}, z$, the bound on stability follows.

\bibliographystyle{IEEEtran}


\end{document}